\documentclass[letterpaper, 10 pt, conference]{ieeeconf}
\IEEEoverridecommandlockouts            
\let\labelindent\relax                  

\pdfoutput=1
\usepackage[utf8]{inputenc}
\usepackage[T1]{fontenc}
\usepackage{graphicx} 
\usepackage{amsmath,amsfonts,amssymb,booktabs,siunitx,leftidx} 
\usepackage{fontawesome}
\usepackage[nolist]{acronym}  
\usepackage[overload]{textcase}
\usepackage[inline]{enumitem} 
\usepackage{flushend} 
\usepackage{pgfplots} 
\pgfplotsset{compat=1.8}
\usepackage{multirow}
\usepackage{threeparttable}
\usepackage{array}
\usepackage{cite}
\usepackage{mathtools}
\usepackage{subcaption}
\usepackage[hidelinks]{hyperref} 
\usepackage{silence}
\newcommand{\figref}[1]{\hyperref[#1]{Fig.~\ref*{#1}}}
\newcommand{\tabref}[1]{\hyperref[#1]{Tab.~\ref*{#1}}}
\newcommand{\secref}[1]{\hyperref[#1]{Sec.~\ref*{#1}}}

\def\ie{\textit{i.e.},}
\def\eg{\textit{e.g.},}
\def\Eg{\textit{E.g.},}
\def\etal{\textit{et al.}}

\def\figvspace{\vspace{-1.2em}}
\newacro{vlm}[VLM]{visual-language model}
\newacro{llm}[LLM]{large language model}
\newacro{3dsg}[3DSG]{3D scene graph}
\newacro{rag}[RAG]{retrieval-augmented generation}
\newacro{slam}[SLAM]{simultaneous localization and mapping}
\newacro{egg}[EGG]{event-grounding graph}

\title{\LARGE \bf Event-Grounding Graph: Unified Spatio-Temporal Scene Graph from Robotic Observations}

\author{Phuoc~Nguyen, Francesco~Verdoja, Ville~Kyrki%
	\thanks{This work was supported by the Research Council of Finland (decision 354909). The authors acknowledge the use of the MIDAS infrastructure of Aalto School of Electrical Engineering. P.\ Nguyen, F.\ Verdoja, and V.\ Kyrki are with School of Electrical Engineering, Aalto University, Espoo, Finland. {\tt\small \{firstname.lastname\}@aalto.fi}}}
\begin{document}
\bstctlcite{IEEEexample:BSTcontrol}
\makeatletter
\let\@oldmaketitle\@maketitle
\renewcommand{\@maketitle}{\@oldmaketitle
	\setcounter{figure}{0}
	\vspace{1em}
	\centering
	\includegraphics[width=\linewidth]{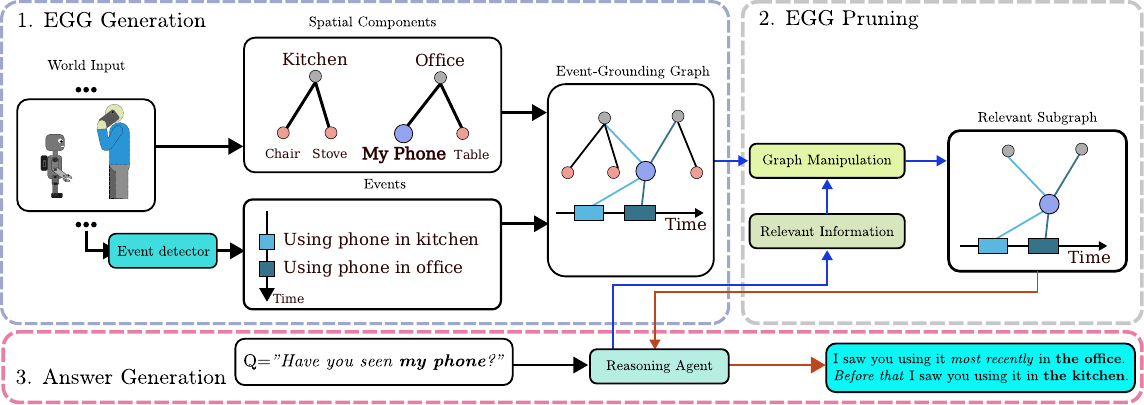}
	\captionof{figure}{\label{fig:cover}We propose~\Acf{egg}, a scene representation framework that grounds observed events with involved spatial elements (\eg{} objects) to allow robots to respond to complex queries.}
    \figvspace}
\makeatother

\maketitle

\begin{abstract}
	A fundamental aspect for building intelligent autonomous robots that can assist humans in their daily lives is the construction of rich environmental representations. While advances in semantic scene representations have enriched robotic scene understanding, current approaches lack a connection between spatial features and dynamic events; \eg{} connecting \textit{the blue mug} to the event \textit{washing a mug}. In this work, we introduce~\acf{egg}, a framework grounding event interactions to spatial features of a scene. This representation allows robots to perceive, reason, and respond to complex spatio-temporal queries. Experiments using real robotic data demonstrate~\ac{egg}'s capability to retrieve relevant information and respond accurately to human inquiries concerning the environment and events within. Furthermore, the~\ac{egg} framework's source code and evaluation dataset are released as open-source at: \url{https://github.com/aalto-intelligent-robotics/EGG}.
\end{abstract}
\acresetall

\section{Introduction}

Memory representations that encapsulate both the geometric and semantic aspects of the environment are crucial for robots to perceive and understand their surroundings while performing physical tasks to assist us in our daily lives. For instance, to answer a question \textit{"Where is my phone?"}, a robot needs to complete three steps: identify objects that could be classified as "phones," determine which one is specifically "my phone," and then locate it. Humans typically recall an object's position based on previous experiences, such as where they \textit{last} used it or where they \textit{commonly} use it.

There has been a surge in research aimed at embedding semantic features into spatial scene elements. Most notable are the development of \acp{3dsg}~\cite{armeni3DSceneGraph2019, guConceptGraphsOpenVocabulary3D2024a, hughesHydraRealtimeSpatial2022c, hughesFoundationsSpatialPerception2024, werbyHierarchicalOpenVocabulary3D2024c, honerkampLanguageGroundedDynamicScene2024b}. By representing spatial concepts such as objects, freespace, and rooms as nodes and their inter-relationships as edges,~\acp{3dsg} boast rich semantic features while remaining compact in memory footprint. While this body of work incorporates semantic information into geometric features of the representation, \eg{} grounding the concept of "my phone" into a node, they do not capture dynamic events, \ie{} interactions that result in changes to the scene, frequently occurring within those environments.

Another line of research focuses on recording and verbalizing robot experiences and observations by leveraging the capabilities of~\acp{vlm}~\cite{anwarReMEmbRBuildingReasoning2025, barmannDeepEpisodicMemory2021, xieEmbodiedRAGGeneralNonparametric2025}. This line of research enhances robots' understanding of how the scene can be interacted with and enables them to respond to free-form spatio-temporal questions from humans. However, these works often describe the environment using free-form natural language captions, which often leads to a lack or misrepresentation of the spatial features of the environment, \eg{} multiple events of "using a phone" can be captured as text captions, but they do not provide information about which "phone" the person is using.

Currently, there is a gap in the representation of the connection between the scene's spatial feature changes and the interactions that cause them. Connecting the semantic attributes of these interactions (\ie{} their symbolic knowledge) with the scene's spatial features is an extension of the grounding problem~\cite{hughesFoundationsSpatialPerception2024, ranaSayPlanGroundingLarge2023c}. Grounding dynamic events through spatial features allows robots to not only perceive the environment, but also to recall and reason from historical interactions, \eg{} offering insights about unusual occurrences, assessing the states of routine tasks, or tracking personal items based on usage history.

In this work, we propose a novel framework,~\ac{egg}, to aggregate, organize, and describe robotic observations over time, creating a unified representation that includes both spatial features and dynamic events. The rich information from the framework efficiently enables the extraction of relevant information subsets based on specific queries or tasks, ensuring insightful and contextually appropriate responses from robots, as illustrated in \figref{fig:cover}. We then demonstrate the capability of our representation through a series of experiments using real robotic data. The core contributions of this work are:
\begin{itemize}
	\item \ac{egg}, a framework that unifies spatial features from the environment with dynamic events, allowing robots to perceive, reason, and explain what happens in a scene.
	\item A method that exploits the spatio-temporal queryability of~\ac{egg} to retrieve a task-related compressed subgraph.
	\item Experimental results from real robotic data showing that~\ac{egg} allows robots to retrieve relevant information and respond to human queries regarding the scene.
	\item An open-source implementation of~\ac{egg} framework as well as the evaluation dataset, promoting further advancements in spatio-temporal scene understanding.
\end{itemize}

\section{Related works}

Enriching geometric robotic maps with semantic information is a vital step towards achieving deeper scene understanding. In addition to methods that integrate rich semantic data into conventional 2D and 3D maps~\cite{rosinolKimeraSLAMSpatial2021, huangVisualLanguageMaps2023a, schmidPanopticMultiTSDFsFlexible2022, mccormacFusionVolumetricObjectLevel2018a, xuMIDFusionOctreebasedObjectLevel2019},~\acp{3dsg}~\cite{armeni3DSceneGraph2019, rosinol3DDynamicScene2020} have emerged as an alternative representation and gathered considerable research interests.~\Acp{3dsg} represents spatial concepts, such as rooms and objects, as nodes. As outlined in~\cite{hughesFoundationsSpatialPerception2024}, 3DSGs are metric-semantic; they aim to ground the semantic information of a scene into the map's geometry, specifically through nodes and edges. Thus, they provide robots with semantic insights about the scene, enabling them to perform advanced tasks autonomously while maintaining a low memory footprint~\cite{hughesHydraRealtimeSpatial2022c, guConceptGraphsOpenVocabulary3D2024a}. Notably, Hughes \etal{} introduces Hydra~\cite{hughesHydraRealtimeSpatial2022c, hughesFoundationsSpatialPerception2024}, a real-time~\ac{3dsg} construction framework that solidifies the practical application of this representation on real robots. ConceptGraphs~\cite{guConceptGraphsOpenVocabulary3D2024a} and HOV-SG~\cite{werbyHierarchicalOpenVocabulary3D2024c} enrich the semantic information available for object nodes in a~\ac{3dsg} by incorporating open-vocabulary features into the frameworks. By leveraging the rich semantic information available in scene graphs, multiple works have utilized reasoning agents such as~\acp{llm} to perform complex tasks such as pick-and-drop~\cite{ranaSayPlanGroundingLarge2023c}, embodied question-answering~\cite{pmlr-v305-saxena25a}, interactive object search~\cite{honerkampLanguageGroundedDynamicScene2024b}, or rearrangements~\cite{mohammadiMOREMobileManipulation2025}. However, current~\acp{3dsg} are limited to grounding semantic information into static spatial concepts and can only handle tasks that do not require awareness of external interactions within the environment. Current spatio-temporal~\ac{3dsg} frameworks, such as~\cite{nguyenREACTRealtimeEfficient2025a, schmidKhronosUnifiedApproach2024c, geDynamicGSGDynamic3D2025a}, only focus on updating the geometric features of the environment, and not on the semantic context that causes scene changes.

Another significant body of research focuses on developing memory representations that incorporate interactions with the environment. B{\"a}rmann \etal{}~\cite{barmannEpisodicMemoryVerbalization2025} propose a framework that systematically records a robot's interaction history within a hierarchical tree-like structure. This representation includes raw robotic sensory inputs at its base layer and abstracted descriptions of the robot's experiences at higher layers. Closer to our work, ReMEmbR~\cite{anwarReMEmbRBuildingReasoning2025} constructs a spatio-temporal representation that aggregates video captions generated by a~\ac{vlm}, along with timestamps and the robot's positional data, into a vector database. Another relevant work is Embodied-RAG~\cite{xieEmbodiedRAGGeneralNonparametric2025}, which develops a spatio-temporal representation in the form of a semantic forest. This representation is constructed by embedding video captions captured at various points in time into the nodes of a topological graph, with a~\ac{llm} summarizer generating higher-level nodes from these embeddings. While these approaches acknowledge the importance of interactions with the environment, they exhibit limitations concerning the grounding of recorded events in the spatial features of the environment. This leads to issues such as multi-view inconsistency and the unnecessary accumulation of redundant data, exemplified by the repetitive description of the same scene upon revisit. Our work addresses this gap.

\section{Problem Formulation}
\label{sec:prob_form}

Formally, a scene can be described by the set of $N$ tracked spatial elements $S$, \eg{} objects, rooms, or free space. Each spatial element $S^{i} \in S$ with $i \in [1, N]$ is characterized by a set of attributes. The choice of attributes is task dependent, but commonly includes at least position in a known world frame and a semantic class or name. We classify attributes into two types: time-invariant, \eg{} semantic class, material, and color, and time-variant, \eg{} position and state of operation. Thus, $S^{i} = ( \mathcal{I}^{i}, \mathcal{V}^{i} )$, with $\mathcal{I}^{i}$  and $\mathcal{V}^{i}$ being the set of time-invariant and time-variant attributes respectively. We denote with $\mathcal{V}^{i}_{T} = \{ \mathcal{V}^{i}_{j}\ |\ t_{j} \in T \}$ the history of time-variant attributes of spatial element $S^{i}$ in the time interval $T$, \ie{} the list of all recorded values of the attributes at any instant within the interval.

We define an event that occurs in the time interval $T$ as a sequence of interactions executed by either the observer or an external agent within the scene, each resulting in an attribute change of at least one tracked spatial element at a particular instant $t \in T$. We denote an event by $E_T = (O_{T}, D_{T}, s_{T})$. $O_{T} \subseteq S$ is the set of spatial elements that experience attribute changes during the event. Attribute changes, $\mathcal{V}_{T}=\{\mathcal{V}_{T}^{i}\ |\ S^{i} \in O_{T} \}$, may include changes to objects' positions, \eg{} someone moving the chairs, the appearance or disappearance of objects, \eg{} someone taking a mug outside of the robot's view, or modifications in their operational states due to interactions, \eg{} someone turning the lights on or off. $D_{T}$ is the set of element-wise descriptions of each change, providing insight into the role that each spatial element plays in the event. Formally, for every attribute change $\mathcal{V}_{T}^{i} \in \mathcal{V}_{T}$, a corresponding description $d^{i}_{T} \in D_{T}$ exists that verbally describes the change. $s_{T}$ is the event summary, which provides an overall description of the event. Collectively, an event $E_T$ not only outlines the scene change, but also describes how each spatial element is affected during the interval $T$.

Our first objective is to create a representation $\mathcal{G}$ that aggregates, describes, and grounds observed events $E$ with the spatial elements $S$ of the scene. Secondly, by harnessing the information from $\mathcal{G}$, we extract task-relevant details to respond to free-form natural language queries $Q$. Particularly, our goal is to predict an answer $A$ given a query $Q$ and a representation $\mathcal{G}$.

\section{Methodology}
\label{sec:method}
\begin{figure}
	\centering
	\includegraphics[width=0.49\textwidth]{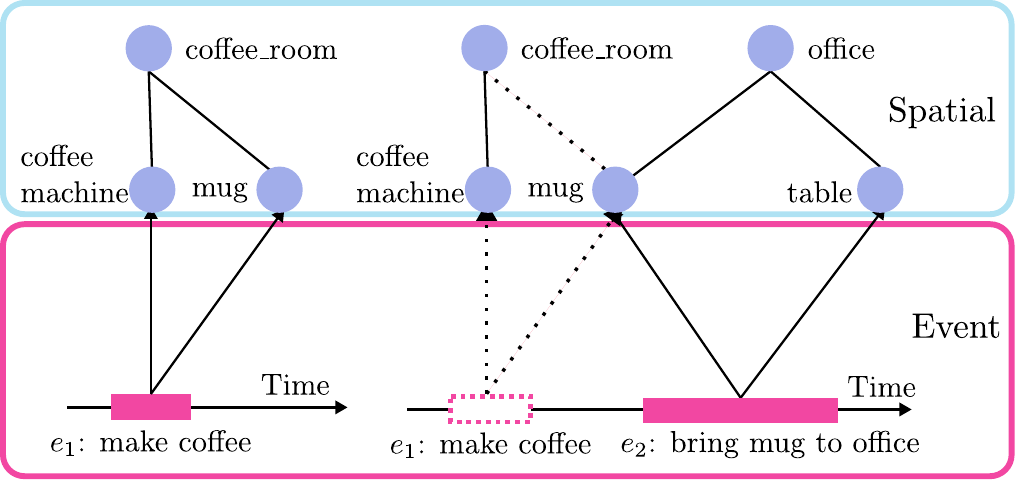}
	\caption{Spatial and event components connections. Spatial and event edges are not permanent and are characterized by time intervals.}
	\label{fig:egg}
	\figvspace{}
\end{figure}

\figref{fig:cover} illustrates the pipeline of our framework. Initially, from observed events, we generate an~\ac{egg} $\mathcal{G}$. Then, using information relevant to a query $Q$, we extract a subgraph $\mathcal{G}_{Q}$. Finally, we send $\mathcal{G}_{Q}$ to a reasoning agent to generate the answer $A$.

\subsection{\ac{egg} generation}
\label{subsec:egg}
\ac{egg} consists of two interrelated sets of components: spatial components and event components, as illustrated in \figref{fig:egg}. Formally,~\ac{egg} is the graph $\mathcal{G} = ( \mathcal{N}^{S}, \mathcal{N}^{E}, \mathcal{E}^{S}, \mathcal{E}^{E} )$, with $\mathcal{N}^{S}$ and $\mathcal{N}^{E}$ being the spatial and event nodes, and $\mathcal{E}^{S}$ and $\mathcal{E}^{E}$ being the spatial and event edges.

The spatial components $\mathcal{G}^{S} = ( \mathcal{N}^{S}, \mathcal{\mathcal{E}}^{S} )$ follow a hierarchical~\ac{3dsg} structure analogous to that of Hydra~\cite{hughesHydraRealtimeSpatial2022c,hughesFoundationsSpatialPerception2024}. Each spatial node $n^{S}_i \in \mathcal{N}^{S}$ corresponds to the attribute set of a spatial element, so $n^{S}_i = S^{i}$. In this work, the constructed~\ac{3dsg} consists of two layers: rooms and objects. However, the formulation from \secref{sec:prob_form} does not prevent the inclusion of other spatial concepts, such as buildings and free space~\cite{hughesHydraRealtimeSpatial2022c,hughesFoundationsSpatialPerception2024} or asset objects~\cite{geDynamicGSGDynamic3D2025a}. Unlike static scene graph frameworks such as~\cite{hughesHydraRealtimeSpatial2022c, hughesFoundationsSpatialPerception2024, guConceptGraphsOpenVocabulary3D2024a}, objects in~\ac{egg} also include time variant attributes, which track changes in objects resulting from event interactions. Each spatial edge $e^{S}_{il} = (T_{il}, n^{S}_i, n^{S}_{l} )$ connecting two nodes, $n^{S}_{i}$ and $n^{S}_{l}$, represents spatial relationships such as inclusion, \eg{} the mug is in the kitchen. Since nodes are not permanent in~\ac{egg}, edges are additionally characterized by the time interval $T_{il}$ during which the spatial relationship exists.

The event components $\mathcal{G}^{E} = ( \mathcal{N}^{E}, \mathcal{E}^{E} )$ represent the observed events and their relationships with the interacted objects. \figref{fig:egg} illustrates how events are grounded with multiple objects. To each event $E_T$ corresponds an event node $n^{E}_{T} = (p_{T}, s_{T})$ defined spatially by the observation points of the robot $p_{T}$, \ie{} the sensor's pose, and temporally by the respective time interval $T$. The event overall description is provided by the summary $s_{T}$. Each event edge $e^{E}_{Ti} = (T, n^{E}_{T}, n^{S}_{i}, d^{i}_{T})$ connects the event $E_{T}$ to the object $S^{i} \in O_{T}$. Similar to the spatial edges, the event edges are also timestamped and exist only in the interval $T$ when the event occurs. The event description $d^{i}_{T} \in D_{T}$ describes the role of the object $S^{i}$ in the event.

\subsection{\ac{egg} pruning}
\label{subsec:spatial_temporal_information_retrieval}

\begin{figure}
	\centering
	\includegraphics[width=0.45\textwidth]{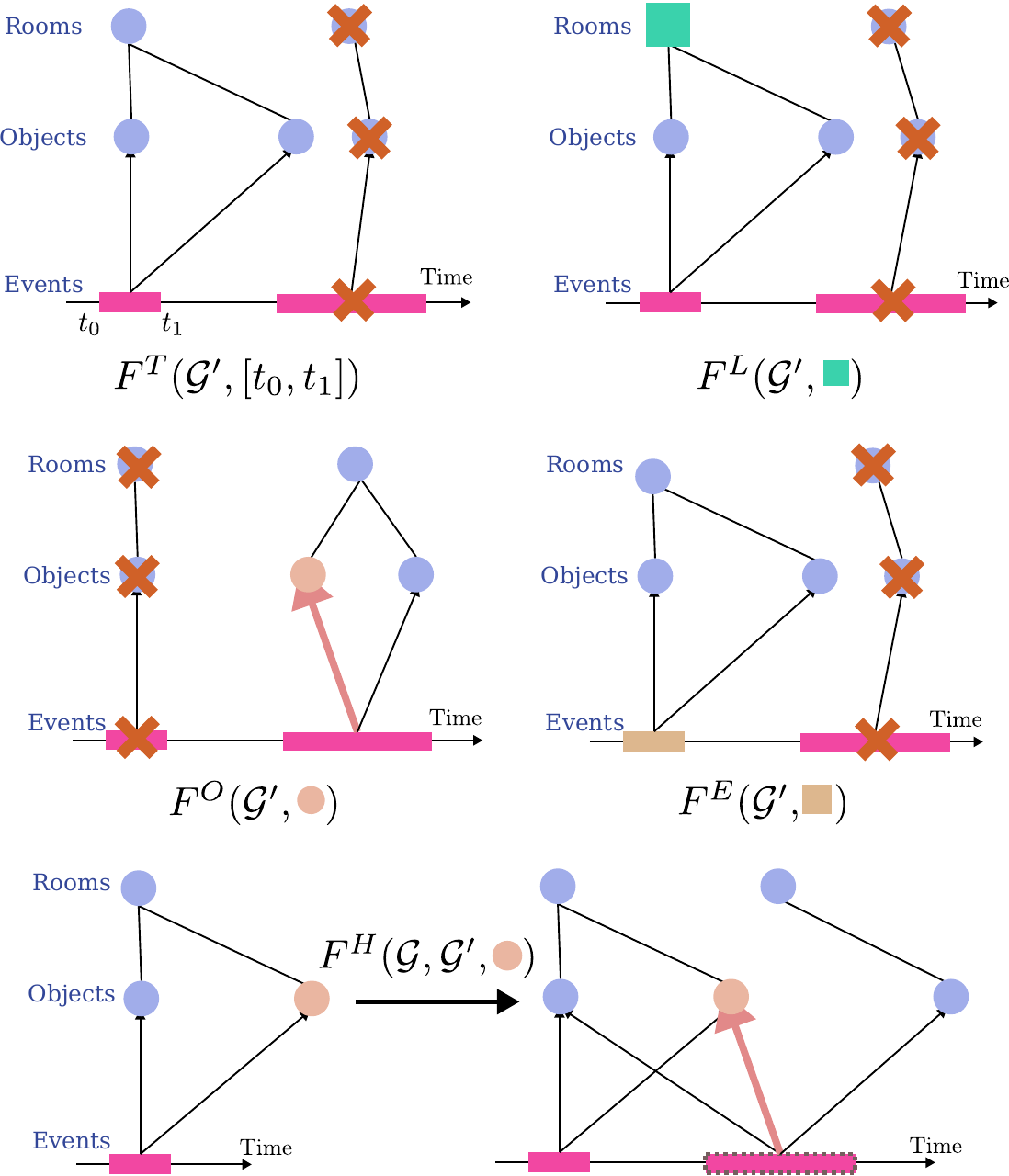}
	\caption{Task-based graph manipulation}
	\label{fig:prune}
	\figvspace{}
\end{figure}

After receiving a query $Q$, the full graph $\mathcal{G}$ is pruned into a subgraph $\mathcal{G}_{Q}$ which contains information relevant to that query. The pruning aims to both decrease the computational effort needed to generate the answer and to improve the quality of the answer by removing potentially distracting irrelevant input from the answer generation process.

To this end, we assume the information relevant to a query, denoted as $I_{Q}$, can be split into four types: time interval $T_{Q}$, locations $L_{Q}$, objects $O_{Q}$, and events $E_{Q}$. We will describe how to retrieve $I_Q$ from $Q$ in \secref{subsec:pruning}. Given the set of relevant information $I_{Q} = ( T_{Q},  L_{Q}, O_{Q}, E_{Q})$ and an~\ac{egg} $\mathcal{G}$, we retrieve the query-relevant subgraph $\mathcal{G}_{Q} \subseteq \mathcal{G}$ by using a series of graph manipulation functions, illustrated in \figref{fig:prune} and detailed below. Each function takes as input a subgraph $\mathcal{G}' = \{ \mathcal{N}^{S'}, \mathcal{E}^{S'}, \mathcal{N}^{E'}, \mathcal{E}^{E'} \} \subseteq \mathcal{G}$ and a subset of $I_Q$ to return another subgraph $\mathcal{G}^{*} = \{ \mathcal{N}^{S*}, \mathcal{E}^{S*}, \mathcal{N}^{E*}, \mathcal{E}^{E*} \} \subseteq \mathcal{G}$. This is achieved first by selecting the list of relevant nodes $\{ \mathcal{N}^{S*}, \mathcal{N}^{E*} \}$ according to a set of rules, and then by adding all edges connected to the selected nodes to $\mathcal{G}^*$, \ie{} $\mathcal{E}^{S*} = \{ e^{S}_{ij}\ \vert \  n^{S}_{i}, n^{S}_{j} \in \mathcal{N}^{S*} \}$ and $\mathcal{E}^{E*} = \{ e^{E}_{Ti},\ \vert \ n^{S}_{i} \in \mathcal{N}^{S*}, n^{E}_{T} \in \mathcal{N}^{E*} \}$. Specifically, the graph manipulation functions are:
\begin{itemize}
	\item \textbf{Time pruning} $F^{T}(\mathcal{G}', T_{Q})$: From a time interval $T_{Q}$, we can select a subgraph $\mathcal{G}^*$ consisting of the set of events that transpire during it. The resulting subgraph consists of event nodes $\mathcal{N}^{E*}=\{ n^{E}_{T}\in \mathcal{N}^{E'} \vert \ T \subseteq T_{Q}\}$. The preserved spatial components are those with at least one connection to the events in the pertinent time window, that is, $\mathcal{N}^{S*} = \{ n^{S}_{i}\in\mathcal{N}^{S'}\ \vert\ (\exists e^{E}_{Ti}\in \mathcal{E}^{E'})\ T \subseteq T_{Q} \}$.
	\item \textbf{Location pruning} $F^{L}(\mathcal{G}', L_{Q})$: From a list of locations $L_{Q}$, we select the subgraph $\mathcal{G}^*$ that consists of events and objects from those locations. Specifically for this work, the locations list $L_{Q} \in \mathcal{N}^{S'}$ is essentially a list of spatial nodes that exists on the room layer. Thus $\mathcal{N}^{S*} = L_Q \cup \{ n^{S}_{i}\in\mathcal{N}^{S'}\ \vert \ (\exists n^{S}_{l} \in L_{Q})\ e^{S}_{li} \in \mathcal{E}^{S'}\}$, and $\mathcal{N}^{E*} = \{ n^{E}_{T}\in\mathcal{N}^{E'}\ \vert \ (\exists n^{S}_{l} \in L_{Q})\ e^{E}_{Tl} \in \mathcal{E}^{E'} \}$.
	\item \textbf{Object pruning} $F^{O}(\mathcal{G}', O_{Q})$: From a list of query-relevant object nodes $O_{Q} \subseteq \mathcal{N}^{S'}$, we select the subgraph $\mathcal{G}^*$ that contains relevant objects and the events that involve them. Specifically, we prune the graph such that the event nodes are connected to the relevant objects, $\mathcal{N}^{E*} = \{n^{E}_{T} \in \mathcal{N}^{E'}\ \vert \ (\exists n^{S}_{i} \in O_{Q})\ e^{E}_{Ti} \in \mathcal{E}^{E'}\}$. Then, the preserved spatial nodes are the ones connected to at least one of those events, \ie{} $\mathcal{N}^{S*} = \{ n^{S}_{i}\in \mathcal{N}^{S'}\ \vert \ (\exists n^{E}_{T} \in \mathcal{N}^{E*})\ e^{E}_{Ti} \in \mathcal{E}^{E'} \}$.
	\item \textbf{Event pruning} $F^{E}(\mathcal{G}', E_{Q})$: From a list of query-relevant event nodes $E_{Q} \subseteq \mathcal{N}^{E'}$, we select the subgraph $\mathcal{G}^*$ that contains relevant events and the objects that they involve. The event nodes in this case are just $\mathcal{N}^{E*} = E_{Q}$, and the spatial nodes are those that are connected to at least one of these events, \ie{} $\mathcal{N}^{S*} = \{ n^{S}_{i} \in \mathcal{N}^{S'}\ \vert \ (\exists n^{E}_{T} \in E_{Q})\ e^{E}_{Ti} \in \mathcal{E}^{E'} \}$.
	\item \textbf{History expansion} $F^{H}(\mathcal{G},\mathcal{G}', O_{Q})$: From a list of objects $O_{Q} \subseteq \mathcal{N}^{S'}$, we expand the graph to include the history of events that involve them. The event nodes that involve $O_{Q}$ are $\mathcal{N}^{E*} = \{ n^{E}_{T} \in \mathcal{N}^{E} \ \vert \ (\exists n^{S}_{i} \in O_{Q})\ e^{E}_{Ti} \in \mathcal{E}^{E}\}$. The spatial nodes include those that are given by $O_{Q}$, and related to the expanded event history, \ie{} $\mathcal{N}^{S*} =  O_{Q} \cup \{ n^{S}_{i} \in \mathcal{N}^{S} \ \vert\ (\exists n^{E}_{T} \in \mathcal{N}^{E*})\ e^{E}_{Ti} \in \mathcal{E}^{E}\}$.
\end{itemize}
Thus, we retrieve the subgraph:
\begin{align}
	\label{eq:pruning}
	\mathcal{G}_{Q} = \mathcal{C}(F^{T}, F^{L}, F^{O}, F^{E}, F^{H})(\mathcal{G}, I_{Q})
\end{align}
where $\mathcal{C}(\cdot)$ is any combination of the graph manipulation functions. We will present our implemented combination in \secref{subsec:pruning}.

\section{Implementation}

In addition to the formulation presented in \secref{sec:method}, our implementation incorporates the following assumptions:

\begin{itemize}
	\item All objects recorded within the events are considered unique entities.
	\item A localization method is available to determine the robot's pose during the mapping sessions~\cite{macenskiSLAMToolboxSLAM2021}.
	\item A method exists to identify moments when human activities occur within the scene and to select the objects relevant to these activities. In the experiments, we manually identify these moments and objects for each robot mapping session.
	\item A method for re-identifying objects post-mapping session is established~\cite{nguyenREACTRealtimeEfficient2025a}. In the experiments, we use ground truth information to re-identify the objects.
\end{itemize}

\subsection{Building \texorpdfstring{\ac{egg}}{} from robotic observations}
\label{subsec:egg_implement}

We construct~\ac{egg} from a stream of multimodal robotic data, including RGB-D images accompanied by frame-wise timestamps and the robot's pose. We derive object data from an RGB-D video stream to generate spatial nodes. In our implementation, the time-variant attribute set $\mathcal{V}^{i} $ for each object node includes the varying positions of the objects and the state of operation. The position of each object is retrieved by projecting its segmented point cloud from the depth image. While we do not explicitly capture the state of operation, they can be inferred from the instance-wise descriptions $D_{T}$ of the event edges, which we will discuss in more detail later in this section. For simplicity, we record only the initial and final positions of each object for each recorded event. To enrich the object nodes and differentiate between instances of the same semantic class, we gather views of each object from multiple perspectives and utilize a~\ac{vlm} (GPT4o~\cite{openai2024GPT4oSystemCard2024}) to generate concise captions that describe their appearance. Thus, the time-invariant attribute set $\mathcal{I}^{i}$ of each object node consists of its name, semantic class, and caption. The room nodes only have time-invariant attributes, which are their manually labeled names and their positions.

To form the event components $\{\mathcal{N}^{E}, \mathcal{E}^{E}\}$ we retrieve event captions using the video captioning model VideoRefer~\cite{yuanVideoReferSuiteAdvancing2025b}. In order to retrieve the summary caption $s_{T}$, we prompt the model to describe the actions of the person in the video, alongside a list of manually selected involved objects $O_{T}$. Furthermore, to generate element-wise descriptions $D_{T}$, we ask the model to express the relationship between the person featured in the video and the corresponding objects. The position $p_{T}$ of the event node is the mean of all the robot's camera positions during the observation time interval $T$. To evaluate the quality of automated video captioning, for the experiments presented in \secref{sec:evaluation}, we also manually generate all corresponding ground-truth event captions.

\subsection{Task-based graph pruning}
\label{subsec:pruning}

The entire~\ac{egg} can be text-serialized into a JSON data format, which can subsequently be parsed directly by a pre-trained~\ac{llm}~\cite{openai2024GPT4oSystemCard2024}. In this implementation, we deploy a multi-stage sampling strategy to approximate $I_Q$, along with the graph manipulation functions in \secref{subsec:spatial_temporal_information_retrieval} to estimate the task-relevant subgraph $\mathcal{G}_{Q}$.
In the first stage, based on the query, we prompt the~\ac{llm} agent to select the relevant time interval $T_{Q}$ and rooms $L_{Q}$ to look for the information. We then retrieve the subgraph:
\begin{align}
	\mathcal{G}_{Q}^{(1)} = F^L(F^T(\mathcal{G}, T_{Q}), L_{Q})
\end{align}
Next, we provide the~\ac{llm} agent with a list of object names and captions and event summary descriptions from $\mathcal{G}_{Q}^{(1)}$. The agent then, based on the semantic context of the query, is tasked to extract the objects $O_{Q}$ and events $E_{Q}$ related to the query. If there are both objects and events related to the query, we merge them to identify the objects that are most relevant to the query and expand their history. More precisely, if $O_{Q} \neq \emptyset \wedge E_{Q} \neq \emptyset$, then $O^*_{Q} = \{ n^{S}_{i} \in O_{Q}\ \vert \ (\exists n^{E}_{T} \in E_{Q})\ e^{E}_{Ti} \in \mathcal{E}^{E(1)} \}$ and $E^*_{Q} = \{ n^{E}_{T} \in E_{Q}\ \vert \ (\exists n^{S}_{i} \in O_{Q})\ e^{E}_{Ti} \in \mathcal{E}^{E(1)} \}$, with $\mathcal{E}^{E(1)}$ being the set of event edges of $\mathcal{G}_{Q}^{(1)}$. \Eg{} if $Q=$\textit{"Where is the mug that I was drinking coffee with in the coffee room yesterday?"}, $O_{Q} = \{\text{all mugs found in the coffee room yesterday}\}$, and $E_{Q} = \{ \text{all events involving a person drinking coffee}\}$. After merging, $O^{*}_{Q}$ will contain only the mugs used for drinking coffee in the coffee room yesterday. In the case that $O_{Q} = \emptyset$ or $E_{Q} = \emptyset$, $O_{Q}$ and $E_{Q}$ are not merged. Thus, we retrieve the subgraph:
\begin{equation}
	\begin{aligned}[b]
		\mathcal{G}_{Q}^{(2)}
		=
		\begin{cases}
			F^{E}(F^{O}(\mathcal{G}_{Q}^{(1)}, O^*_{Q}), E^*_{Q}), & \text{if } O_{Q} \neq \emptyset \wedge E_{Q} \neq \emptyset \\
			F^{O}(\mathcal{G}_{Q}^{(1)}, O_{Q}),                   & \text{if } O_{Q} \neq \emptyset \wedge E_{Q} = \emptyset    \\
			F^{E}(\mathcal{G}_{Q}^{(1)}, E_{Q}),                   & \text{if } O_{Q} = \emptyset \wedge E_{Q} \neq \emptyset    \\
			\emptyset,                                             & \text{if } O_{Q} = \emptyset \wedge E_{Q} = \emptyset       \\
		\end{cases}
	\end{aligned}
\end{equation}
Finally, if $O_{Q} \neq \emptyset$, we retrieve the task-relevant subgraph $\mathcal{G}_{Q}$ by expanding the event history of relevant objects and pruning the graph to the relevant time interval $T_{Q}$. Intuitively, we expand the history of the relevant objects to retrieve events involving them at all locations in~\ac{egg}, since objects could move around and events that happen before the key events in $E_{Q}$ could have a role in determining the answer for $Q$. We then prune the graph to the relevant time interval to reduce the graph size and remove the events that are too far in the past or future.
\begin{equation}
	\begin{aligned}[b]
		\mathcal{G}_{Q} & =
		\begin{cases}
			F^{T}(F^{H}(\mathcal{G}, \mathcal{G}_{Q}^{(2)}, O^*_{Q}), T_{Q}), & \text{if } O_{Q} \neq \emptyset \wedge E_Q \neq \emptyset \\
			F^{T}(F^{H}(\mathcal{G}, \mathcal{G}_{Q}^{(2)}, O_{Q}), T_{Q}),   & \text{if } O_{Q} \neq \emptyset \wedge E_Q = \emptyset    \\
			\mathcal{G}_{Q}^{(2)},                                            & \text{if } O_{Q} = \emptyset                              \\
		\end{cases}
	\end{aligned}
\end{equation}

Finally, we text-serialize $\mathcal{G}_{Q}$ into JSON format, which is then integrated into the prompt for the~\ac{llm} agent to respond to the user's query\footnote{All prompts and specific~\ac{llm} used can be found in the paper's codebase.}.

\section{Evaluation}
\label{sec:evaluation}

\subsection{Experimental setup}
\label{subsec:exp_setup}

To evaluate~\ac{egg}, we set up and conduct a series of experiments using the representation to perform an information retrieval task, where an~\ac{llm} agent uses the information from the scene to answer queries in natural language. With these experiments, we aim to address the following research questions:

\begin{itemize}
	\item What is the impact of the spatial and event components of~\ac{egg} on the performance of information retrieval tasks, both individually and in combination?
	\item How does pruning~\ac{egg} influence information retrieval accuracy and token usage?
	\item In what ways does the accuracy of generated event captioning affect scene representation, and what is its impact on downstream information retrieval tasks?
\end{itemize}

Currently, to the best of our knowledge, there is no suitable dataset to evaluate~\ac{egg}'s capability. We acknowledge the existence of prior datasets that document human activities in natural settings~\cite{heilbronActivityNetLargescaleVideo2015, zhouAutomaticLearningProcedures2018}. However, these datasets are primarily designed for video captioning research and do not include crucial robotic data, such as depth images and poses, which are necessary for building and evaluating~\ac{egg}. Another notable example is the Ego4D dataset, which contains over 3000 hours of daily-life activity videos captured from an egocentric perspective, along with object detection labels and 3D poses for some samples~\cite{graumanEgo4DWorld30002022}. Yet, the videos in this dataset are not recorded within a consistent environment, nor do they provide instances where the same object is reused, rendering it inadequate for benchmarking~\ac{egg}'s capability to maintain a history of each object.

To address this gap, we evaluate our framework using real-world data collected from a Hello Robot Stretch 2 mobile manipulator platform with an Astra 2 RGB-D camera mounted on top. Data collection was conducted in two locations within our department building: a coffee room and an office. Our dataset includes 35 videos, captured by the robot, of an individual performing various daily activities and interacting with 21 distinct objects in a coffee room and an office. On average, an event consists of 3 to 4 objects being interacted with. Additionally, we provide frame-wise timestamps, object instance segmentation, event-relevant object identification, and local positions of the robot.

The robot task involves verbalizing its observations of a scene through different forms of scene representation. We manually label each video with ground truth summary captions and object role captions to assess the capability of~\acp{vlm} in comparison to human judgment. For each event, we also manually provide segmentation masks for each object at their first and last appearances, enabling projection to capture their instance positions and incorporate them into the position history of their respective spatial nodes (refer to \secref{subsec:egg_implement} for implementation details of the spatial graph). The dataset comprises 80 questions with ground truth answers across four modalities, \textit{for instance}\footnote{The full dataset, including the events and question-answer pairs, will be made publicly available.}:
\begin{itemize}
	\item \textit{text} (24 queries): Returns an answer in natural language. \Eg{} \textit{Which room was the phone last seen used in?}
	\item \textit{binary} (28): Returns True or False. \Eg{} \textit{Was the red ceramic mug ever used to make coffee?}
	\item \textit{node} (18): Returns a list of spatial nodes. \Eg{} \textit{The person left something on the office table at some point between 15:30:00 and 16:16:00. What was it?}
	\item \textit{time} (10): Returns a point in time. \Eg{} \textit{What is the earliest time the person was seen working?}
\end{itemize}
To assess the accuracy of the generated responses for the queries with \textit{text} modality we employ~\ac{llm} semantic scores, where we employ a~\ac{llm} to judge each generated answer with a score from 0 to 1. This evaluation method was chosen over manual evaluation to mitigate potential human biases in judgment and leverage the established capability of~\acp{llm} to provide evaluation scores that consider context accuracy and nuance~\cite{liLLMsasJudgesComprehensiveSurvey2024}. For the \textit{time} modality, we consider an answer to be correct if it lies within 2 minutes of the ground truth answer. Additionally, we provide quantifiable F1 scores for \textit{node} ($F1^{node}$) and \textit{binary} ($F1^{binary}$) modalities.

In our experiments, GPT4o~\cite{openai2024GPT4oSystemCard2024} is utilized to handle object captioning, graph pruning, and reasoning, and VideoRefer~\cite{yuanVideoReferSuiteAdvancing2025b} is utilized for automatic video captioning.

\begin{table}
	\centering
	\resizebox{0.49\textwidth}{!}{
		\begin{threeparttable}
			\caption{\label{tab:qa_eval}Evaluating the capability of~\ac{egg} in verbalizing robot observations against baseline methods. The results are averaged over 5 trials. VideoRefer~\cite{yuanVideoReferSuiteAdvancing2025b} is used for automatic captioning of events for \ac{egg} and ReMEmbR.}
			\small

			\begin{tabular}{lcccccc}
				\toprule
				\textbf{Method}                                                                                                            & $\mathbf{S}^{\mathbf{text}}\uparrow$ & $\mathbf{F1}^{\mathbf{binary}}\uparrow$ & $\mathbf{F1}^{\mathbf{node}}\uparrow$ & $\mathbf{A}^{\mathbf{time}}\uparrow$ \\
				\midrule
				\textit{\ac{egg}}                                                                                                          & \textbf{0.70}                        & \textbf{0.78}                           & \textbf{0.76}                         & \textbf{0.78}                        \\
				\textit{Hydra\tnote{\textasteriskcentered}~\cite{hughesHydraRealtimeSpatial2022c, hughesFoundationsSpatialPerception2024}} & 0.52                                 & 0.00                                    & 0.68                                  & 0.10                                 \\
				\textit{ReMEmbR}\tnote{\textdagger}~\cite{anwarReMEmbRBuildingReasoning2025}                                               & 0.54                                 & 0.49                                    & \faTimes                              & 0.38                                 \\
				\bottomrule
			\end{tabular}
			\begin{tablenotes}\footnotesize
				\item[\textasteriskcentered] Implemented by removing the event nodes $\mathcal{N}^{E}$ and event edges $\mathcal{E}^{E}$ from \ac{egg}
				\item[\textdagger] Since ReMEmbR does not have access to the scene graph, we did not evaluate it on the queries with \textit{node} modality.
			\end{tablenotes}

		\end{threeparttable}
	}
	\figvspace{}
\end{table}

\subsection{Experimental results}
\label{subsec:experiments}

We evaluate our method against two baselines, a purely spatial memory representation, Hydra~\cite{hughesHydraRealtimeSpatial2022c,hughesFoundationsSpatialPerception2024}, and a spatio-temporal memory building method, ReMEmbR~\cite{anwarReMEmbRBuildingReasoning2025}. Our experimental results, presented in \tabref{tab:qa_eval}, clearly demonstrate that our method outperforms both baselines across all modalities. From these findings, it can be inferred that the unified spatio-temporal features of~\ac{egg} preserve the advantages of both purely spatial and temporal representations, enabling the~\ac{llm} agent to reason more effectively, especially on instance-based queries.

Qualitatively, Hydra~\cite{hughesHydraRealtimeSpatial2022c, hughesFoundationsSpatialPerception2024}, a purely spatial scene representation, struggles to address event-related queries. By observing the reasoning output, the~\ac{llm} often rely solely on the positions of objects to infer their relationships. However, this approach is ineffective for queries that require detailed information about human interactions, which are not typically captured in a~\ac{3dsg}. For instance, the query \textit{"Describe the appearance of the most frequently used mug for making coffee at the coffee machine"} requires usage frequency of different mugs to correctly answer.

On the other hand, ReMEmbR~\cite{anwarReMEmbRBuildingReasoning2025} performs better on descriptive queries with its access to event data. Yet, it lacks the grounding of events with spatial features, so the~\ac{llm} can only rely on video captions. Notably, it fails to address queries such as \textit{"How many different mugs were seen in the events?"}, which could be easily answered with spatial representations. Furthermore, it struggles with queries requiring the usage history of specific objects, such as \textit{"Describe the appearance of the most frequently used mug for making coffee at the coffee machine"}.

Overall, it is evident that each component of~\ac{egg}---the spatial, event components---plays a crucial role in enriching the information provided to the~\ac{llm} agent.

\subsection{Ablation study}
\begin{table*}
	\centering
	\caption{\label{tab:qa_ablation}Ablation study on \ac{egg}. The results are averaged over 5 trials.\tnote{\textasteriskcentered}}
	\small
	\resizebox{\textwidth}{!}{
		\begin{threeparttable}

			\begin{tabular}{lccccccc}
				\toprule
				\textbf{Method}                   & $\mathbf{A}^{\mathbf{all}}\uparrow$ & $\mathbf{S}^{\mathbf{text}}\uparrow$ & $\mathbf{F1}^{\mathbf{binary}}\uparrow$ & $\mathbf{F1}^{\mathbf{node}}\uparrow$ & $\mathbf{S}^{\mathbf{time}}\uparrow$ & \textbf{\text{Tokens (1K)}} $\downarrow$ & \textbf{Compression} $\uparrow$ \\
				\midrule
				\textit{\ac{egg}}                 & \textbf{0.83}                       & \textbf{0.85}                        & \textbf{0.77}                           & \textbf{0.87}                         & 0.80                                 & 998                                      & 64.72 \%                        \\
				\textit{\ac{egg} (w.o edges)}     & 0.79                                & 0.81                                 & 0.72                                    & 0.82                                  & 0.80                                 & \textbf{922}                             & \textbf{68.02} \%               \\
				\textit{\ac{egg} (w.o pruning)}   & 0.77                                & 0.80                                 & 0.65                                    & 0.79                                  & \textbf{0.88}                        & 1576                                     & 0.00  \%                        \\
				\textit{\ac{egg} (auto-captions)} & 0.76                                & 0.70                                 & 0.78                                    & 0.76                                  & 0.78                                 & 985                                      & 66.55 \%                        \\

				\bottomrule
			\end{tabular}
			\begin{tablenotes}\footnotesize
				\item[\textasteriskcentered] Ground truth captioning is used for event nodes in all versions except for \textit{auto-captions}, for which VideoRefer~\cite{yuanVideoReferSuiteAdvancing2025b} is used.
			\end{tablenotes}

		\end{threeparttable}
	}
	\figvspace{}
\end{table*}

The results from \tabref{tab:qa_ablation} illustrate several ablation studies on different versions of \ac{egg}. In addition to the metrics mentioned in \secref{subsec:exp_setup}, we provide an additional accuracy score across all modalities ($A^{all}$), a tokens count and graph compression percentage, which is the ratio of the number of characters of the JSON-serialized subgraph $\mathcal{G}_{Q}$ with the JSON-serialized full graph $\mathcal{G}$.

\subsubsection{The role of event edges}

To assess the importance of event edges, we perform pruning on the original~\ac{egg} and omit them when JSON-serializing the graph (\textit{w.o edges} in \tabref{tab:qa_ablation}). This version of~\ac{egg} performs close to the standard version, but misses some needed details for understanding specific interactions. In most cases, the pruning process essentially removes unrelated objects of the same class from the resulting graph. \Eg{} for the query $Q_{1}=$\textit{"Which mug was most often used for coffee-making?"}, only the mugs involved in the coffee-making events are present in the subgraph $\mathcal{G}_{Q_1}$. Although complete edge information is valuable and the standard version shows higher performance, removing edges after pruning can still provide reasonable performance when token limits are a consideration.

\subsubsection{Graph pruning efficiency}

To assess how the subgraph $\mathcal{G}_{Q}$ affects reasoning performance and token usage, we compare two versions of the information search strategy: one utilizing the pruning method detailed in \secref{subsec:pruning} and the other using the complete graph with the~\ac{llm} agent (\ac{egg} w.o pruning in \tabref{tab:qa_ablation}). The version employing graph pruning outperforms the alternative and achieves approximately 65\% graph compression, resulting in a 36\% reduction in token usage. It should be noted that the token count includes the input prompt tokens. The pruned version of~\ac{egg} demonstrated significantly higher overall semantic score in the downstream information retrieval task. Disregarding the excess input tokens, we also hypothesize that this improvement may stem from the~\ac{llm}’s tendency to hallucinate when presented with redundant information, as discussed in studies like \cite{shiLargeLanguageModels2023a}. Moreover,~\acp{llm} have shown positional biases towards information presented at the beginning or end of the input prompt~\cite{xiaoEfficientStreamingLanguage2024b}. The experimental results indicate that the pruning process improves~\ac{llm} performance on downstream tasks by excluding task-irrelevant entities from the representation.

\subsubsection{Video captioning capability}

In this study, we construct~\ac{egg} using two different video captioning methods: using ground truth labeling and using a~\ac{vlm}, VideoRefer~\cite{yuanVideoReferSuiteAdvancing2025b}. As anticipated, the quality of video captioning directly affects the accuracy of subsequent tasks. As detailed in~\tabref{tab:qa_eval}, automatic captioning reduces the overall accuracy by 9\% compared to using ground truth event labeling. These results underscore the necessity for advancements in video captioning models.

\subsection{Failure modes analysis}
\label{subsec:failure_modes}

\begin{figure}[!htb]
	\centering
	\includegraphics[width=0.43\textwidth]{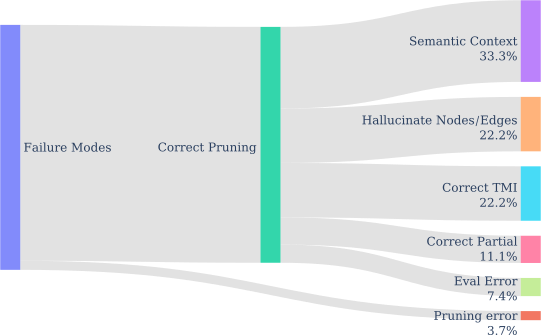}
	\caption{Summary of failure modes of~\ac{egg} with pruning and ground truth captioning.}
	\label{fig:failure_modes}
	\figvspace{}
\end{figure}

\figref{fig:failure_modes} provides a summary of the potential failure modes identified during an experimental trial using ground truth event captions (shown in \tabref{tab:qa_ablation}). Our observations show that most errors originate from the~\ac{llm} answer generation. More specifically, the primary sources of error include:
\begin{itemize}
	\item Semantic Context: the~\ac{llm} misinterprets the question or the context of an event, leading to irrelevant answers.
	\item Hallucinate Nodes and Edges: the~\ac{llm} occasionally makes incorrect connections between nodes that do not exist in the graph or misidentifies objects and events related to the query.
\end{itemize}
In contrast, we identified only 1 instance where pruning removed a vital node related to the query, suggesting that~\ac{egg} provides adequate information for reasoning tasks and effectively supports graph-based reasoning with current~\acp{llm}, and the shortcomings mostly arise from the interaction with the~\ac{llm}.

To increase performance, incorporating more specialized models or integrating graph reasoning tools could help the~\ac{llm} better interpret the structured information provided by~\ac{egg}, thereby reducing errors and improving understanding in complex tasks.

\section{Discussions}

Our experiments demonstrate that~\ac{egg}'s grounding of observed dynamic events with spatial features allows it to retain the capabilities of purely spatial and temporal representations while improving understanding of historical interactions within a scene. We have also shown that incorporating a graph pruning strategy to select relevant information improves the performance on the downstream information retrieval task and reduces token usage. By retaining both geometric changes and semantic context,~\ac{egg} could provide valuable information for robots to execute higher level tasks, such as making more informed decisions during semantic object search~\cite{guConceptGraphsOpenVocabulary3D2024a, honerkampLanguageGroundedDynamicScene2024b}, or learning mobile manipulation skills from human demonstrations~\cite{merloExploitingInformationTheory2025a}.

However, a notable constraint of our framework is that the subgraph $\mathcal{G}_{Q}$ as formulated in~\eqref{eq:pruning} is not guaranteed to be optimal. We observe that often the number of relevant objects typically returned can be redundant, as our heuristic approach may include irrelevant objects connected to pertinent events. This redundancy also applies to the events. Thus, this underscores the necessity for the development of more robust semantic search algorithms, especially when extending to larger, more complex environment with more events. Another constraint of~\ac{egg} lies in the lack of connectivity between the recorded events. As a result, the~\ac{llm} agent has to infer from scattered events and form implicit links based on the timestamps or event ordering. A hierarchical structure in the time domain, such as having summaries of multiple events~\cite{barmannEpisodicMemoryVerbalization2025}, could further improve the performance of our framework. Moreover, the process of building~\ac{egg} is currently offline, making it not yet suitable for real-time robotic applications. Currently, our method requires that any information queried be already present in the map, a limitation shared by many semantic mapping techniques. This issue is however further compounded by the addition of event memory. Efforts to overcome this limitation, such as those that focus on inferring unseen interactions and transitions \cite{schmidKhronosUnifiedApproach2024c}, are crucial and hold great promise for long-term robotic operation.

\section{Conclusion}

In this work, we present~\ac{egg}, a novel scene graph framework that unifies the benefits of spatial and temporal scene representations. We achieve this by grounding observed dynamic events to the scene's spatial features, resulting in a richer description of the scene that allows an off-the-shelf~\ac{llm} to answer complex queries about the scene. We evaluated our method by constructing~\ac{egg} from real-world robotic data, and tested its capability on a downstream information retrieval task. Our experimental results indicate that each component of~\ac{egg} contributes to enriching~\ac{egg}'s information and allows the~\ac{llm} to respond correctly to more questions. Furthermore, the experiments have shown evidence that the spatio-temporal queryability of~\ac{egg} combined with a graph pruning strategy improves the result on the downstream task by removing task-irrelevant information. 

Capabilities like multi-robot coordination, robot teaching by demonstration, and natural human-robot cooperation depend on the ability of robots to operate in dynamic human-inhabited environments. Real-time spatio-temporal representations supporting comprehension of physical entities, interactions between them, and other agents' actions will be instrumental in solving those tasks and deploying intelligent industrial and service robots.

\section{Acknowledgement}

The authors acknowledge the use of Grammarly's generative AI tool and OpenAI's GPT4o~\cite{openai2024GPT4oSystemCard2024} for improving readability across all sections. Additionally, the authors acknowledge the role of OpenAI's GPT4o~\cite {openai2024GPT4oSystemCard2024} in evaluating our method as detailed in \secref{sec:evaluation}, as well as generating documentation and improving code readability in the project's code base.
\bibliographystyle{IEEEtran}
\bibliography{ref}

\end{document}